%% file: main.tex
\documentclass[a4paper]{article}
\usepackage{INTERSPEECH2019,amssymb,amsmath,epsfig}
\usepackage{microtype,url}
\usepackage{graphicx}
\usepackage{multirow}
\usepackage{booktabs}
\usepackage{amssymb,amsmath,graphicx}
\usepackage{caption}
\usepackage{subcaption}
\usepackage[export]{adjustbox}
\usepackage{moredefs} 
\usepackage{color}

\newcommand{\red}[1]{{\color{red} \textbf{#1}}}
\newcommand{\green}[1]{{\color{green} \textbf{#1}}}
\defcommand{\vec}[1]{\mathbf{#1}} 

\def\vx{\mathbf{x}}

\def\vy{\mathbf{y}}

\def\ve{\mathbf{e}}
\def\vL{\mathbf{L}}

\title{Two-Pass End-to-End Speech Recognition}
\name{Tara N. Sainath$^*$\thanks{$^*$~Equal contribution.}, Ruoming Pang$^*$\footnotemark[1],
  David Rybach, Yanzhang He, Rohit Prabhavalkar, Wei Li, \\ Mirk\'{o} Visontai,
  Qiao Liang, Trevor Strohman, Yonghui Wu, Ian McGraw, Chung-Cheng Chiu}
\address{Google, Inc., USA}
\email{\{tsainath,rpang\}@google.com}

\begin{document}
\maketitle
\ninept
\input{abstract}
\input{intro}
\input{2pass}
\input{experiments}

\input{results}
\input{conclusions}
\bibliographystyle{IEEEbib}
\newpage
\bibliography{main}
\end{document}

%% file: abstract.tex
\begin{abstract}

The requirements for many applications of state-of-the-art speech recognition
systems include not only low word error rate (WER) but also low latency.
Specifically, for many use-cases, the system must be able to decode utterances
in a streaming fashion and faster than real-time.
Recently, a streaming recurrent neural network transducer (RNN-T) end-to-end
(E2E) model has shown to be a good candidate for on-device speech recognition,
with improved WER and latency metrics compared to conventional on-device models
\cite{Ryan19}.
However, this model still lags behind a large state-of-the-art conventional model in
quality~\cite{Golan16}.
On the other hand, a non-streaming E2E Listen, Attend and Spell (LAS) model has
shown comparable quality to large conventional models~\cite{CC18}.
This work aims to bring the quality of an E2E streaming model closer to
that of a conventional system by incorporating a LAS network as a second-pass
component, while still abiding by latency constraints.
Our proposed two-pass model achieves a 17\%-22\% relative reduction in
WER compared to RNN-T alone and increases latency by a small fraction over RNN-T.
\end{abstract}

%% file: intro.tex
\section{Introduction \label{sec:introduction}}

There continues to be a growing popularity with end-to-end models (E2E) for speech recognition~\cite{Ryan19,CC18,Graves12, GravesMohamedHinton13,RaoSakPrabhavalkar17,Chan15, KimHoriWatanabe17,ChiuRaffel17}.
These models, which fold the acoustic, pronunciation and language models (AM, PM, LMs) into a single network, have shown competitive results compared to conventional ASR systems which have separate AM, PM, and LMs.
E2E models are particularly attractive for on-device ASR, as they can outperform
on-device conventional models \cite{McGraw16} of comparable size.

Running ASR on-device with direct user interaction, presents numerous challenges.
First, the recognition results must be streaming. That is, words should appear on the screen as soon as they are spoken. Second, the model must have a small latency (i.e., the delay between the user speaking and the text appearing ), thus running at or faster than real-time on mobile devices. Third, the model must be able to utilize user context \cite{Petar15} (e.g., list of contacts, song names, etc.) to improve recognition accuray. Recently, we presented a RNN-T E2E model that satisfies these constraints~\cite{Ryan19}.
However, the RNN-T model's quality still lags behind that of a large conventional model \cite{Golan16}.

Non-streaming E2E models, such as Listen, Attend and Spell (LAS) \cite{Chan15}, have shown competitive performance to a large conventional model \cite{CC18}. However, LAS models are not streaming as they must attend to the entire audio segment, making it challenging to use them in interactive applications.

In two-pass decoding, the second pass model is often used to improve the initial outputs from first-pass models by using lattice rescoring~\cite{Ortmanns1997} or n-best reranking~\cite{Schwartz1991}. Keeping user-perceived latency low while obtaining the quality gains is the main challenge with applying second-pass models. Language model rescoring is commonly been used for multi-pass decoding~\cite{sundermeyer2015,liu2016,kumar2017,Anjuli18}, but more recently has been used with a LAS model to rescore hypotheses from a first-pass conventional model~\cite{RohitAnal17}. We can think of LAS decoder, which takes acoustic information from the encoder and language model information from previous predictions, as being strictly stronger than second-pass language models. Thus, in this work, we explore using the LAS model for second-pass processing.

Specifically, we explore a two-pass architecture in which an RNN-T decoder and a LAS decoder share an encoder network. Sharing the encoder allows us to reduce model size and computation cost compared with having dedicated encoders for RNN-T and LAS.
During inference, the RNN-T model produces streaming predictions while in the end the LAS decoder finalizes the prediction. We explore tradeoffs by running the LAS decoder as a beam search versus rescoring hypotheses from RNN-T.

Our experiments are conducted on a $\sim$30,000 hour voice
search task. We find that with LAS second-pass beam search,
we can get a 15\% relative improvement over first-pass RNN-T
for a shorter utterance (SU) test set, but the model degrades
on longer utterances (LU), a common problem for
attention models~\cite{Jan15}. In contrast, second-pass
rescoring gives us a much better tradeoff for SU and LU compared
to beam-search. Next, we experiment with ways to improve the
rescoring model WER by changing the training objective function
to more closely match rescoring. Specifically, we apply a
minimum word error rate (MWER) training strategy \cite{prabhavalkar2018minimum}
where hypotheses from RNN-T are used as inputs to the LAS decoder
and the LAS decoder is trained to minimize expected word error
rates. In addition, we reduce computation cost by running the
first-pass RNN-T model with an adaptive beam \cite{Lowerre76}
and pruning the first-pass lattice before rescoring. Overall,
we find that our proposed LAS rescoring model provides 17\% to 22\%
relative improvement in WER compared to a first-pass
RNN-T model, without a degradation in biasing accuracy. In
addition, the second-pass LAS decoder increases finalization latency by
less than 200ms, which has been considered the limit of acceptable
interactive latency \cite{Miller1968}.

The rest of this paper is organized as follows. Section \ref{sec:2pass} describes the two-pass architecture and various inference strategies explored in this paper. Experiments are presented in Section \ref{sec:experiments} while results are discussed in Section \ref{sec:results}. Finally, Section \ref{sec:conclusions} concludes the paper and discusses future work.

%% file: 2pass.tex
\section{Two-Pass E2E ASR \label{sec:2pass}}

\subsection{Model Architecture}

\begin{figure}[h!]
  \centering
  \includegraphics[scale=0.4]{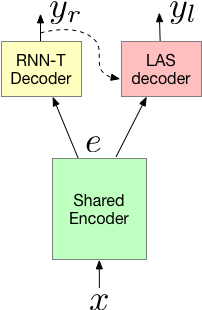}
  \caption{{Two-Pass Architecture}}
   \label{fig:2pass_architecture}
   \vspace{-0.2in}
\end{figure}

The proposed two-pass architecture is shown in Figure \ref{fig:2pass_architecture}.
We denote the parameterized input acoustic frames as $\vx=(\vx_1 \ldots \vx_T)$,
where $\vx_t \in \mathbb{R}^d$ are log-mel filterbank energies in
this work ($d=80$) and $T$ denotes the number of frames in $\vx$.
In the first pass, each acoustic frame $\vx_t$ is passed through a shared encoder,
consisting of a multi-layer LSTM, to get output $\ve_t$, which is passed to an RNN-T
decoder for producing $\vy_r$ at each time step in a streaming fashion.
In the second pass, the output of the shared encoder of all frames
$\ve=(\ve_1 \ldots \ve_T)$ is passed to a LAS decoder.
During training, the LAS decoder computes output $\vy_l$ according to $\ve$. During decoding it may additionally use $\vy_r$ as described below.

\subsection{Decoding \label{sec:2pass-decoding}}

We explore using the LAS decoder in two different decoding modes in this work. Specifically,
\begin{itemize}
\item In the ``2nd beam search'' mode, it produces output $\vy_l$ from $\ve$
  alone, ignoring $\vy_r$, the output of the RNN-T decoder.
\item In the ``rescoring'' mode, we first pick the top-K hypotheses from the RNN-T decoder.
We then run the LAS decoder on each sequence in the teacher-forcing mode, with
attention on $\ve$, to compute a score, which combines log probability of the
sequence and the attention coverage penalty~\cite{Jan17}.
The sequence with the highest LAS score is picked as the output sequence.
\end{itemize}

\subsection{Training}

In this section, we describe the training stategies for the two-pass model.

\subsubsection{Combined Loss}

In theory we can train a two-pass model from random initialization with the
following combined loss, where $\vy^*$ represents the ground truth transcript:
\begin{equation}
  \vL_{\texttt{combined}}(\vx, \vy^*) = \lambda \vL_{\texttt{RNNT}}(\vx, \vy^*) + (1 - \lambda) \vL_{\texttt{LAS}}(\vx, \vy^*)
  \label{eq:2pass-loss}
\end{equation}
In the above equation, $\lambda$ is a hyperparameter, which we set to be 0.5 in our setup to equally weight the RNN-T and LAS losses.
In practice we find training directly from scratch to be unstable, mainly because the losses for RNN-T and LAS are in drastically different ranges when training from scratch. Therefore, we take a multi-step process to train the model:
\begin{enumerate}
\item Train an RNN-T model as in~\cite{Ryan19};
\item Take the encoder trained in step (1), freeze it, and train a LAS decoder
as in~\cite{CC18}.
\item ``Deep finetuning'': train the shared encoder and both decoders at the same time with the combined loss.
\end{enumerate}

\subsubsection{MWER training}

One of the drawbacks of the loss in Equation \ref{eq:2pass-loss} is that the second-pass LAS decoder is optimized independently of the RNN-T decoder. This means that there is a mismatch between the training and decoding strategies outlined in Section \ref{sec:2pass-decoding}.

To address this, we use an additional training step to further refine the LAS decoder to minimize errors, following the MWER training process introduced in~\cite{prabhavalkar2018minimum}. Specifically, given input $\vx$, groundtruth transcript $\vy^*$,
the probability computed by LAS $P(\vy_m|\vx)$ for any given target sequence $\vy_m$ with
teacher-forcing (where $m=r$ if $\vy_m$ is given by RNN-T and $m=l$ if $\vy_m$ is given by LAS), we refine the pre-trained two-pass model as follows.

First, we run a beam search with one of the decoders $m$ from the two-pass model to get a set of hypotheses $H_m=\{h_1, \ldots, h_b\}$ where $b$ is the beam-size.
To make the MWER training match decoding, the generation of $H_m$ depends on the
target decoding mode.
For a LAS decoder to be used in the ``2nd beam search'' mode, we compute $H_m$ by
running beam search with the LAS decoder itself on $\vx$ ($m=l$).
For a LAS decoder
to be used in the ``rescoring'' mode, on the other hand, we compute $H_m(\vx)$ by
running beam search with the first-pass RNN-T decoder ($m=r$).

For each sequence $\vy_m \in H_m$, let $W(\vy^*, \vy_m)$ be the number of word errors of
$\vy_m$, let $\overline{W}(\vy^*, H_m) = \frac{1}{|H_m|}\sum_{\vy_m \in H_m}{W(\vy^*,
  \vy_m)}$ be the mean number of word errors for $H_m$, and let $\hat{W}(\vy^*, \vy_m) =
W(\vy^*, \vy_m) - \overline{W}(\vy^*, H_m)$ be the relative word error rate of
$\vy_m$ in $H_m$.
We also let $\hat{P}(\vy_m|\vx, H_m) = \frac{P(\vy_m|\vx)}{\sum_{\vy_i \in
    H_m}P(\vy_i|\vx)}$ represent the conditional probability LAS decoder assigns
to hypothesis $\vy_m$ among all hypotheses in $H_m$. The MWER loss is defined as
\begin{equation}
\vL_{\texttt{MWER}}(\vx, \vy^*) = \sum_{\vy_m \in H_m(\vx)}\hat{P}(\vy_m | \vx, H_m) \hat{W}(\vy^*, \vy_m)
  \label{eq:mwer}
\end{equation}

We train the LAS decoder to minimize a combination of the MWER loss and the
maximum-likelihood cross-entropy loss:
\begin{equation}
  \vL_{\texttt{MWER}}(\vx, \vy^*) + \lambda_{\texttt{MLE}}\log{P(\vy^*|\vx)}
  \label{eq:mwer_ce}
\end{equation}
where $\lambda_{\texttt{MLE}}$ is a hyperparameter that experimentally we set to be $\lambda_{\texttt{MLE}} = 0.01$ following \cite{prabhavalkar2018minimum}.

%% file: experiments.tex
\section{Experimental Details \label{sec:experiments}}

\subsection{Data Sets}
Our experiments are conducted on a $\sim$30,000 hour training set consisting of
43 million English utterances.
The training utterances are anonymized and hand-transcribed, and are
representative of Google's voice search traffic in the United States.
Multi-style training (MTR) data are created by artificially corrupting the clean
utterances using a room simulator, adding varying degrees of noise and
reverberation with an average SNR of 12dB~\cite{Chanwoo17}.
The noise sources are drawn from YouTube and daily life noisy environmental
recordings.
The main test sets we report results on include $\sim$14K short utterances (\emph{SU}) less than 5.5 seconds long and $\sim$16K long utterances (\emph{LU}) greater than 5.5 seconds, both extracted from Google traffic.

To evaluate the performance of contextual biasing, we report
performance on a contacts test set, which consists of requests to call/text contacts. This set is created  by mining contact names from
the web, and synthesizing TTS utterances in each of these categories using
a concatenative TTS approach with one voice~\cite{Gonzalvo16}. Noise is then artificially added to the TTS data, similar to the process
described above \cite{Chanwoo17}.
To bias model predictions towards contacts, we construct a biasing FST on a list
of contact phrases and perform shallow-fusion between the biasing FST and E2E
model during inference.
We refer the reader to \cite{Pundak18} for more details regarding E2E
shallow-fusion biasing.

\subsection{Model Architecture Details}
All experiments use 80-dimensional log-Mel features, computed with a 25ms window
and shifted every 10ms.
Similar to~\cite{Golan16}, at the current frame, $t$, these features are stacked
with 2 frames to the left and downsampled to a 30ms frame rate.

The same encoder network described in~\cite{Ryan19} is used for all experiments.
It consists of 8 LSTM layers, where each layer has 2,048 hidden units followed
by a 640-dimensional projection layer.
We insert a time-reduction layer with the reduction factor $N=2$ after the
second LSTM layer of encoder.

The RNN-T decoder contists of a prediction network and a joint network. The
prediction network has 2 LSTM layers of 2,048
hidden units and a 640-dimensional projection per layer as well as an embedding
layer of 128 units. The outputs of encoder and prediction network are fed to a
joint network that has 640 hidden units. The LAS decoder consists of multi-head
attention~\cite{Vaswani17} with four attention heads, which is fed into 2 LSTM
layers of 2,048 hidden units and a 640-dimensional projection layer. It has an
embedding layer of 96 units. Both decoders are trained to predict 4,096 word
pieces~\cite{Schuster2012}, which are derived using a large corpus of text
transcripts.

The total size of the RNN-T model is 114M parameters, and the additional
second-pass LAS decoder is 33M parameters. All models are trained in
Tensorflow~\cite{AbadiAgarwalBarhamEtAl15} using the Lingvo
\cite{shen2019lingvo} toolkit on $8 \times 8$ Tensor Processing
Units (TPU) slices with a global batch size of 4,096. 

\subsection{Measuring Latency}

As computing devices may vary, we use a simplified model of computation to
estimate latency. First, we assume that the bandwidth $K$ on CPU is 10GB/second; this
number is within the range of modern mobile CPUs. We also make the non-optimal
assumption that each hypothesis is computed independently, meaning that the
major operations are matrix/vector multiplies, the time of which will be
dominated by the speed of loading matrix parameters into the CPU.

Assuming no interrupts or batching across beam search hypotheses, the
latency is calculated from Equation \ref{eq:latency} when doing fixed beam
decoding/rescoring with $H$ hypotheses over $N$ tokens. When using an adaptive beam, where a lattice is generated,
we assume $H \cdot N$ is now replaced by the number of lattice arcs when calculating latency.

\begin{equation}
  \textnormal{latency} = \frac{1}{K} \cdot H \cdot N \cdot M_{\textnormal{decoder}}
\label{eq:latency}
\end{equation}
where $M_{\textnormal{decoder}}$ denotes the number of bytes in the decoder part of the model.

We report latency on the 90\%-tile \emph{LU} set which has longer utterances.
We assume the 90\%-tile contains roughly 295 audio frames and a target
sequence of $N = 28$ tokens. Finally, $M_{\textnormal{decoder}}$ is
33MB, assuming 33M parameters of the LAS decoder which are quantized, which
we have found has a negligible degradation in accuracy \cite{Ryan19}.
Our goal is to ensure that the second-pass latency on the  90\%-tile is under
200ms to that user-perceived latency is minimized \cite{Miller1968}.
%

%% file: results.tex
\section{Results \label{sec:results}}

\subsection{2nd Beam Search}

Table \ref{table:2ndbeam} shows the results running the LAS decoder in the 2nd
beam-search mode.
For comparison, the table also shows two baselines \texttt{B0-B1}, namely an
RNN-T only and a LAS-only model, trained separately from scratch.
All results are obtained with fixed beam-size of $H=8$.

Experiment \texttt{E0} indicates that when the encoder is initialized from an
RNN-T model and held fixed, the LAS decoder performs worse than a LAS-only model
with a dedicated encoder (\texttt{B1}), demonstrating the challenges in sharing
a single encoder with different types of decoders by adapting the LAS decoder
alone.
When we jointly train the encoder and both decoders in a model initialized from
\texttt{E0}, the model quality (\texttt{E1}) improved in both SU and LU over
\texttt{E0}.
Overall we find that 2nd beam search improves over RNN-T (\texttt{B0}) on SU but
degrades on LU, a common issue with attention models for long utterances
\cite{Jan15}.

\begin{table} [h!]
  \centering
  \caption{WER Results, LAS Beam Search.}
  \begin{tabular}{|c||c|c|c|} \hline
    Exp-ID & Model & SU & LU \\ \hline
    \textit{B0} & RNN-T & 6.9 & 4.5 \\ \hline
    \textit{B1} & LAS-only & 5.4 &  4.5  \\ \hline
    \textit{E0} & Frozen Shared Enc & 6.4 & 5.3 \\ \hline
    \textit{E1} & Deep Finetuned & 6.1 & 4.8 \\ \hline
  \end{tabular}
  \label{table:2ndbeam}
  \vspace{-0.1in}
\end{table}

%
%
%


\subsection{Rescoring}


We noticed that the RNN-T-only model (\texttt{B0}) has much lower oracle WERs
than its decoding WERs.
This motivates us to explore rescoring RNN-T hypothesis with the LAS decoder.
Table \ref{table:rescoring} compares the performance of running LAS with beam
search (\texttt{E1}) to with rescoring (\texttt{E2}).
The table shows that rescoring takes a small hit in WER on SU compared to beam
search, probably because the first-pass RNN-T decoder, with a much higher SU WER
of 6.9 (\texttt{B0}), generates a set of hypotheses with slightly lower quality
than those generated by the LAS decoder during beam search.
However, rescoring's quality on LU is much better than that of beam search,
likely because RNN-T (\texttt{B0}) performs much better on longer utterances
compared to LAS.
Overall, LAS rescoring not only improves SU WER significantly upon the first
pass RNN-T, but also improves WER for LU, demonstrating that rescoring is able
to combine the strengths of RNN-T and LAS. Since rescoring gives us the best
tradeoff between quality on SU and LU, we will focus only on rescoring and
present further improvements in the next section.

\begin{table} [h!]
  \centering
  \caption{WER results, LAS Rescoring.}
  \begin{tabular}{|c||c|c|c|} \hline
    Exp-ID & Decoding & SU & LU \\ \hline
    \textit{B0} & RNN-T & 6.9 & 4.5 \\ \hline
    \textit{B1} & LAS-only & 5.4 &  4.5 \\ \hline
    \textit{E1} & Beam Search & \textbf{6.1} & 4.8 \\ \hline
    \textit{E2} & Rescoring & 6.2 & \textbf{4.1} \\ \hline
  \end{tabular}
  \label{table:rescoring}
    \vspace{-0.1in}
\end{table}

%
%
%

%


\subsection{Further Rescoring Improvements}

\subsubsection{Adaptive Beam}

To bridge the gap between two-pass 2nd beam search vs. rescoring on SU, we first
explore increasing the diversity of rescoring candidates with a larger
first-pass RNN-T beam.
Table \ref{table:adaptive} shows that as beam size is increased
(\texttt{E2-E4}), the WER improves, but naturally at cost of proportionally
increased first-pass computation cost.
To address this, we look at an adaptive beam search strategy \cite{Lowerre76}.
Specifically, we prune first-pass beam candidates if they are too far in
threshold from the current best candidate, where the threshold optimizes
first-pass latency following \cite{Ryan19}.
The table shows that with an adaptive beam (\texttt{E5}), we can achieve similar
WER to a fixed but large beam (\texttt{E3}).

\begin{table} [h!]
  \centering
  \caption{Rescoring WER with first-pass fixed vs. adaptive beam.}
  \begin{tabular}{|c||c|c|c|} \hline
    Exp-ID & First-pass Max Beam Size & SU &  LU  \\ \hline
    \textit{E2} & Fixed, 8  & 6.2 & 4.1   \\ \hline
    \textit{E3} & Fixed, 10 & 6.2  & 4.1 \\ \hline
    \textit{E4} & Fixed, 16 & 6.1  & 4.1  \\ \hline \hline
    \textit{E5} & Adaptive, 10  & 6.2  & 4.0  \\ \hline
  \end{tabular}
  \label{table:adaptive}
    \vspace{-0.1in}
\end{table}

The adaptive-beam also has the additional benefit that it generates a lattice to
use for rescoring, rather than an N-best list. Rescoring a lattice is more
efficient than rescoring an N-best list, as it avoids duplicate computation on
the common prefixes between candidate sequences, and thus should reduce latency.
As a reminder, latency in
Equation \ref{eq:latency} is now calculated by looking at the total arcs in the lattice.
Table ~\ref{table:adaptiveops} compares adaptive beam to
a fixed beam with N-best rescoring, where we rescore all first-pass hypotheses.
The table show that with an adaptive beam and lattice rescoring, we can reduce latency
compared to a fixed beam with N-best rescoring. However, the latency is still above our budget.

\begin{table} [h!]
  \centering
  \caption{Latency vs. Rescoring Methods.}
  \begin{tabular}{|c||c|c|c|c|} \hline
  Strategy & Latency (ms)  \\ \hline
  1st-pass Fixed, N-best Rescoring & 369.6      \\ \hline
  1st-pass Adaptive, Lattice Rescoring & \textbf{247.5} \\ \hline
  \end{tabular}
  \label{table:adaptiveops}
    \vspace{-0.1in}
\end{table}

To reduce latency further, we explore reducing the number of maximum arcs in the
lattice rescored at each step. Table ~\ref{table:2ndrescops} shows we can limit the
rescored hypotheses to 4, which we find does not degrade accuracy and also reduces
latency. Overall, the second-pass decoder rescoring an adaptive-beam lattice
fits within our 200ms latency budget.
%

\begin{table} [h!]
  \centering
  \caption{Two-Pass Performance vs. Las Rescoring Beam Size.}
  \begin{tabular}{|c||c|c|c|c|} \hline
    Beam Size & SU & LU & Contacts & Latency (ms)  \\ \hline
    2 & 6.2 & 4.0 & 7.5 & -  \\ \hline
    4 & \textbf{6.2} & \textbf{4.0} & \textbf{7.1} & \textbf{171.6} \\ \hline
    8 & 6.2 & 4.0 & 7.1 & 247.5 \\ \hline
  \end{tabular}
  \label{table:2ndrescops}
    \vspace{-0.2in}
\end{table}

\subsubsection{MWER}

Finally, we report two-pass results after MWER training our model.
Since the LAS decoder will be used for rescoring, we use RNN-T to provide
the candidate hypotheses for LAS decoder MWER training.
Table \ref{table:embr} shows that MWER improves rescoring WER for both SU and LU by
8\% relative. Overall, the two-pass rescoring model gives a 17\% and 22\%
relative reduction in both SU and LU, respectively.

\begin{table} [h!]
  \centering
  \caption{Two-pass rescoring results after MWER training.}
  \begin{tabular}{|c||c|c|c|c|} \hline
    Exp-ID & Model & SU & LU & Contacts \\ \hline
    \textit{B0} & RNN-T only & 6.9 & 4.5 & 7.0 \\ \hline
    \textit{E6} & No MWER & 6.2 & 4.0 & 7.1 \\ \hline
    \textit{E7} & MWER & \textbf{5.7} & \textbf{3.5} & \textbf{7.0} \\ \hline
  \end{tabular}
  \label{table:embr}
    \vspace{-0.1in}
\end{table}

\subsection{Comparison To Large Conventional Model}
A goal of our work is to achieve in an E2E system with comparable performance to
a large conventional model \cite{Golan16}. In this light, we compare the
performance of our proposed two-pass rescoring model to a large conventional
model through a ``side-by-side'' (SxS) evaluation with previously unseen utterances.
In this experiment, each utterance is transcribed by both the conventional and
two-pass models. We collect 500 utterances where the transcription differs
between the two models, and send these utterances to be rated by two human
transcribers. Each transcript is rated as either a win by two-pass over the
conventional model (only two-pass is correct), a loss in two-pass over the
conventional model (only the conventional model is correct), or neutral (both models are correct or incorrect).
Unlike automatic WER evaluations, this side-by-side evaluation allows raters to decide that two different transcripts are both correct; this sometimes leads to different conclusions
than an automatic evaluation would.
We report the following statistics to quantitatively evaluate the SxS:
\begin{itemize}
\item Changed: \% of utterancs in which the two models produced different hypotheses
\item Wins: \# of utts the two-pass hypothesis is correct and conventional model is incorrect
\item Losses: \# of utts the two-pass hypothesis is incorrect and conventional model is correct
\item Neutral: \# of utts the two-pass and conventional model are both correct or incorrect
\item p-Value: Statical significance of WER change with two-pass compared to conventional model
\end{itemize}

Table \ref{table:sxs} shows that the two-pass model changes about 13\% of traffic. The two-pass model has slightly more losses (61) than wins (48) compared to the conventional model, but the majority of the hypotheses have a neutral rating (391) between the two systems. Overall, the p-Value shows the performance difference between the two models is statistically insignificant.

\begin{table}[h!]
  \centering
  \caption{SxS results for Conventional vs. Two-pass}
  \begin{tabular}{|c|c|c|c|c|} \hline
    Changed (\%) & Win & Loss & Neutral & p-Value \\ \hline
    13.2 & 48 & 61 & 391 & 10.0\%-20.0\%  \\ \hline
  \end{tabular}
  \label{table:sxs}
    \vspace{-0.05in}
\end{table}

A further analysis of errors is shown in Table \ref{table:wins_and_losses}. The two-pass model is trained with an order of magnitude less text-only data compared to the conventional model, and thus loses on proper nouns (PN) and also due to a weak language-model (wLM). On the contrary, since the two-pass model is trained in the written domain and learns text normalization (TN) implictly, it wins in this area compared to the conventional model which has a separate rule-based text-norm step.

\begin{table} [h!]
\centering
\caption{Analysis of Errors of Conventional vs. Two-Pass Model.}
\begin{tabular}{|c||c|c|c|} \hline
 & Type & Conventional &  Two-Pass \\ \hline
Loss & PN & \footnotesize{alice's restaurant} & \footnotesize{\red{allison's restaurant}} \\
& wLM & \footnotesize{47-in sony plasma tv} & \footnotesize{47-in sony \red{pricing} tv} \\ \hline
 Win & TN & \footnotesize{www nytimes.com} & \footnotesize{\green{www.nytimes.com}} \\
& TN & \footnotesize{john smiths office} & \footnotesize{john \green{smith's} office} \\ \hline
\end{tabular}
\vspace{-0.15in}
\label{table:wins_and_losses}
\end{table}

%% file: conclusions.tex
\section{Conclusions \label{sec:conclusions}}
In this paper, we present a two-pass E2E solution. Specifically, we use a second-pass LAS decoder to rescore hypotheses from a first-pass RNN-T system. We find that this approach gives a 17\% to 22\% reduction in WER compared to RNN-T only, and increases latency by less than 200ms.